\documentclass[journal,twoside,web]{ieeecolor}
\usepackage{generic}
\usepackage{cite}
\usepackage{amsmath,amssymb,amsfonts}
\usepackage{algorithmic}
\usepackage{graphicx}
\usepackage{algorithm,algorithmic}
\usepackage{hyperref}
\usepackage{textcomp}
\usepackage{acronym}
\usepackage{soul}
\usepackage{algorithm}
\usepackage{algorithmic}
\usepackage{xcolor}
\usepackage{threeparttable}    
\usepackage{multirow}

\usepackage{tablefootnote}

\def\BibTeX{{\rm B\kern-.05em{\sc i\kern-.025em b}\kern-.08em
    T\kern-.1667em\lower.7ex\hbox{E}\kern-.125emX}}
\newacro{ECG}{electrocardiography}
\newacro{SCG}{seismocardiography}
\newacro{IMU}{inertial measurement unit}
\begin{document}
\title{Vib2ECG: A Co-Located Chest-Lead SCG–ECG Dataset and Benchmark for ECG Reconstruction}
%Toward Chest-Lead ECG Reconstruction from Cardiac Mechanical Signals
%\title{A New ECG-IMU Paired Dataset for ECG Estimation from Mechanical Activities of the Heart at Variable Positions}
\author{Guorui Lu, \IEEEmembership{Student Member, IEEE}, Xiaohui Cai, \\ Todor Stefanov, \IEEEmembership{Member, IEEE}, Qinyu Chen, \IEEEmembership{Member, IEEE}
\thanks{The study was approved by the Faculty of Science Ethics Review Committee of Leiden University (No. 2026-014). Written informed consent was obtained from all participants.}
\thanks{Guorui Lu, Todor Stefanov, and Qinyu Chen are with the Leiden Institute of Advanced Computer Science (LIACS), Leiden University, Leiden, The Netherlands. (email: g.lu@liacs.leidenuniv.nl, q.chen@liacs.leidenuniv.nl)}
\thanks{Xiaohui Cai is with the School of Computer Science and Technology, University of Science and Technology of China, Anhui, China. }
}

\maketitle

\begin{abstract}
% The twelve-lead electrocardiography (ECG) is a widely used clinical tool for monitoring and diagnosing cardiovascular diseases. Long-term and stable acquisition of ECG signals in daily life is of great significance for the research, monitoring, prevention, and treatment of cardiac disorders. However, the complexity of ECG acquisition hardware leads to high equipment costs, which limits large-scale and continuous use in everyday settings. 
% Recent studies have explored reconstructing ECG signals from cardiac mechanical activities, as mechanical signals, such as seismocardiography (SCG), can be acquired using low-cost sensors. However, due to the limitations of existing datasets, current methods can only estimate limb-lead ECG signals, whereas the diagnosis of cardiac diseases often requires multiple leads, including chest (chest) leads. 
Twelve-lead electrocardiography (ECG) is essential for cardiovascular diagnosis, but its long-term acquisition in daily life is constrained by complex and costly hardware. Recent efforts have explored reconstructing ECG from low-cost cardiac vibrational signals such as seismocardiography (SCG), with prior studies primarily focusing on limb leads, while clinical diagnosis requires multi-lead ECG, including chest leads.
In this work, we propose Vib2ECG, a dataset of synchronized, co-located ECG and chest-wall vibration recordings~\footnote{The dataset can be found in \url{https://leidenuniv1-my.sharepoint.com/:f:/g/personal/lug_vuw_leidenuniv_nl/IgDg5j2CIATlQ56qjY4YIf-vAanPDRScblrc056wRheo7Us?e=KZEeAS}}, which includes complete twelve-lead ECGs and vibrational signals acquired by inertial measurement units (IMUs) at the six standard precordial electrode sites (V1–V6) from 17 subjects. 
Based on this dataset, we also provide a benchmark. Experimental results demonstrate the feasibility of reconstructing electrical cardiac signals at variable locations from vibrational signals using a lightweight  364\,K-parameter U-Net. Furthermore, we observe a hallucination phenomenon in the model, where ECG waveforms are generated in regions where no corresponding electrical activity is present. We analyze the causes of this phenomenon and propose potential directions for mitigation. 
This study demonstrates the feasibility of mobile-device-friendly ECG monitoring through chest-lead ECG prediction from low-cost vibrational signals acquired using IMU sensors.
It expands the application of cardiac vibrational signals and provides new insights into the spatial relationship between cardiac electrical and mechanical activities with spatial location variation. 
% \textcolor{red}{ SCG should be mentioned? when you talk about mechanical signals.}

%The release of this dataset aims to reduce the cost of ECG monitoring, expand the application value of cardiac mechanical signals, and promote understanding of how the relationship between cardiac electrical and mechanical signals varies with spatial location.

\end{abstract}

\begin{IEEEkeywords}
Dataset, ECG, SCG, ECG Waveform Reconstruction.
% \underline{https://www.ieee.org/publications/services/thesaurus-acce}\\
% \underline{ss-page.com.}
\end{IEEEkeywords}

\section{Introduction}
\label{sec:introduction}

\IEEEPARstart{E}{lectrocardiography} (ECG) is a widely used clinical tool to monitor and diagnose cardiac diseases. In the standard twelve-lead system, there is a mature theoretical framework by which physicians can interpret the ECG waveform and analyze the patient’s condition~\cite{zipes2018braunwald}. However, for chronic cardiovascular diseases that require daily ECG monitoring~\cite{wang2024systematic, jaradat2025non, lazaro2024tracking, zhu2024genes, jons2019clinical, chow2025continuous}, wearable ECG devices such as the Holter monitor face challenges. Due to the small amplitude and susceptibility to interference of ECG signals~\cite{zhang2025opportunities}, additional circuitry is required to reduce noise. This both increases the complexity and the costs of ECG devices~\cite{freund2023old}, thereby limiting large-scale use in everyday life. In addition, the complexity of the ECG device circuit also makes its daily use uncomfortable, for example, the device is relatively large and heavy, and multiple electrodes must be attached to the body~\cite{hyun2024evaluation}.

Seismocardiography (SCG) is an emerging monitoring tool that uses inertial measurement units (IMUs) or accelerometers to record vibration signals generated by the human heart~\cite{sorensen2018definition}, which contain useful information about cardiac function~\cite{yang2017combined, rai2021comprehensive, agam2025evaluation}. 
SCG has received increasing attention in recent years, as its cost of sensing hardware continues to decrease as sensor technology advances~\cite{taebi2019recent}. Currently, IMUs can cost as little as \$1~\cite{liu2022evaluation}, whereas ECG devices often cost over \$149~\cite{kulkarni2021ambulatory}. Moreover, SCG can be acquired using everyday devices such as smartphones~\cite{hossein2024smartphone, albrecht2022seismocardiography, lahdenoja2017atrial} or wristbands~\cite{carek2017seismowatch, cao2022guard}, which are already equipped with the IMU or accelerometers, further reducing deployment cost and improving accessibility.

However, SCG also has its limitations. Due to the relatively late start of research on SCG~\cite{zanetti2013seismocardiography}, the interpretation of SCG waveforms lacks a mature theoretical framework comparable to that of ECG.
As a result, SCG signals cannot be directly interpreted by clinicians.
Moreover, SCG is easily affected by external factors such as motion~\cite{shandhi2019performance, skoric2024wavelet}, which means that although it contains rich information, the signals are difficult to distinguish with the naked eye, further increasing the difficulty of interpretation. Although existing studies have explored the direct application of SCG~\cite{semiz2020non, liu2024camera, ganti2022wearable, kimball2021unifying, chan2022respiratory, elnaggar2021detecting, shandhi2020estimation}, they typically focus on a specific disease or parameter estimation and are unable to provide a comprehensive interpretation of the signals comparable to the theoretical framework of ECG.

%and limiting their adoption in practical healthcare settings~\cite{bhat2024enhancing}.

\begin{figure*}[t]
    \centering
    \includegraphics[width=1\textwidth]{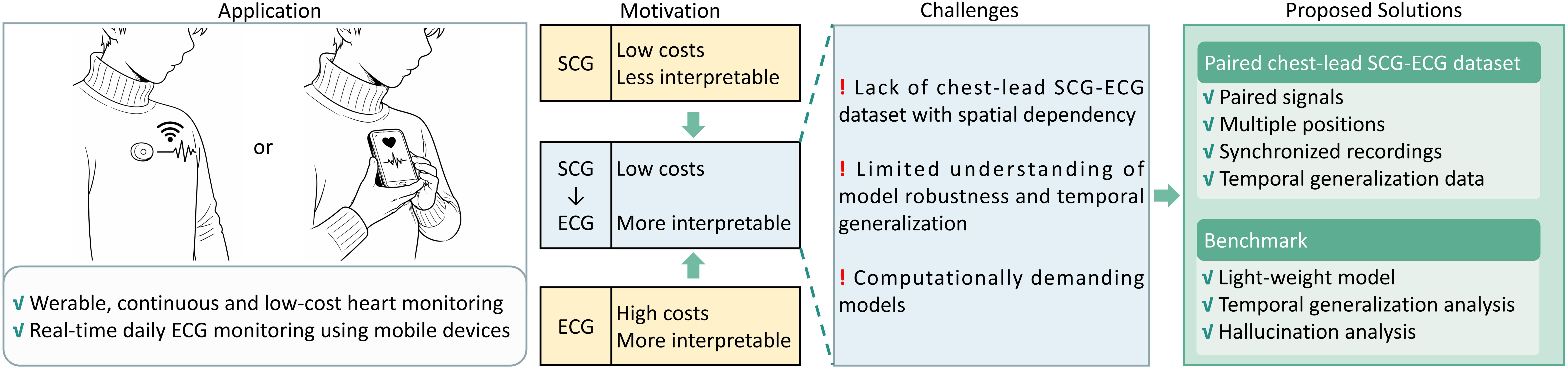}
    \caption{Conceptual overview of SCG-to-ECG reconstruction for mobile and wearable cardiac monitoring. While SCG offers a cost-effective sensing modality, its limited interpretability motivates the reconstruction of ECG signals. This work provides synchronized twelve-lead ECG recordings and vibrational signals acquired with an IMU co-located with each of the six standard precordial electrodes (V1–V6), together with a SCG-to-ECG reconstruction benchmark.}
    \label{fig:concept}
    \vspace{-3mm}
\end{figure*}

To overcome the respective limitations of ECG and SCG while combining their complementary strengths, a promising approach is to acquire SCG signals and then convert them into ECG signals.
As illustrated in Fig.~\ref{fig:concept}, this strategy retains the low-cost benefit of SCG acquisition, while enabling signal interpretation based on the well-established theoretical framework of ECG~\cite{park2020heartquake, cao2022guard, satija2022unified, chen2022contactless}. The physiological motivation behind this approach lies in the electromechanical nature of the heart. 
The heart functions as an electromechanical pump, in which myocardial contraction is initiated by electrical depolarization~\cite{pfeiffer2014biomechanics}. 
As a result, the heart’s rhythmic mechanical activity occurs under the coordination of electrical signals~\cite{prinzen2011mechano}. 
These electrical and mechanical processes are therefore tightly coupled and their corresponding signals exhibit correlation. 
Hence, in theory, it is possible to reconstruct one modality of signal from the other.

However, existing SCG-to-ECG conversion methods still have limitations. 
First, the studies discussed above focus on limb-lead ECG reconstruction, whereas the diagnosis of many cardiovascular diseases requires ECG recordings from multiple anatomical locations, particularly chest leads~\cite{bacevicius2023six, yoon2024classification, zepeda2024detection}. Second, these studies do not provide mechanical recordings co-located with each of the six standard precordial electrodes (V1–V6), limiting systematic investigation of location-dependent SCG–ECG relationships. 
Third, prior methods have not investigated the temporal generalization of learned models over time, leaving their robustness across different time periods largely unexplored.

Datasets with mechanical sensors co-located at the six standard precordial electrode sites (V1–V6) can support systematic investigation of location-dependent SCG–ECG relationships. 
To fill this gap, we release a paired ECG-SCG dataset collected at chest lead positions from 17 subjects. 
This dataset provides synchronized ECG and vibrational recordings with an IMU co-located with each of the six standard precordial electrodes (V1–V6), enabling investigation of how the ECG–SCG relationship varies by sensor placement. 
Based on this dataset, we provide a subject-specific benchmark, propose a reconstruction method that decomposes the vibration signals into low-frequency and high-frequency components for joint ECG reconstruction, and offer a physiological and physical explanation. 
In the process, we identify a “hallucination” phenomenon during ECG reconstruction, in which reconstructed ECG waveforms appear in regions without corresponding electrical activity. 
We analyze its cause and offer potential directions for mitigation. 
Finally, to study the temporal generalization of models in this task, data were collected in daily recording sessions over eight consecutive days from the same subject to explore how a model trained on the first day performs when tested on subsequent days, thereby preliminarily investigating temporal generalization for SCG-to-ECG reconstruction. In summary, our contributions are as follows:
\begin{itemize}
    \item We release Vib2ECG, a dataset containing synchronized twelve-lead ECG and vibrational recordings from 17 subjects, with an IMU co-located with each of the six standard precordial electrodes (V1–V6).

    \item Based on the Vib2ECG dataset, we provide a benchmark model for ECG reconstruction using a subject-specific approach and demonstrate that a lightweight U-Net model (364\,K parameters) can reconstruct ECG signals of chest leads with an average L1 distance of 0.1524.

    \item We systematically identify and analyze a hallucination phenomenon in ECG reconstruction, where fake ECG waveforms are generated without corresponding electrical activity, and investigate its underlying causes.

    \item We provide a preliminary investigation into temporal generalization by evaluating model performance across multi-day recordings from the same subject.

\end{itemize}

% Our ultimate goal is to develop a system that can be deployed on mobile devices, enabling automatic prediction of corresponding chest-lead ECG signals when the device is positioned on the chest. The present study represents the first step toward this long-term goal. It addresses the current lack of paired ECG and vibrational signal data covering chest-lead locations and is expected to facilitate not only the reconstruction of ECG from SCG, but also a wider range of research related to SCG and its physiological relevance.

Our ultimate goal is to develop a mobile-deployable system that can predict chest-lead ECG signals using a single smart patch or smartphone placed on the chest. To enable this, the model must learn to infer the corresponding chest-lead ECG solely from the recorded vibrational signal. This motivates the use of co-located SCG–ECG recordings at the six standard precordial electrode sites (V1–V6). The present study provides such recordings and serves as a first step toward this goal.

\begin{figure*}[t]
    \centering
    \includegraphics[width=\textwidth]{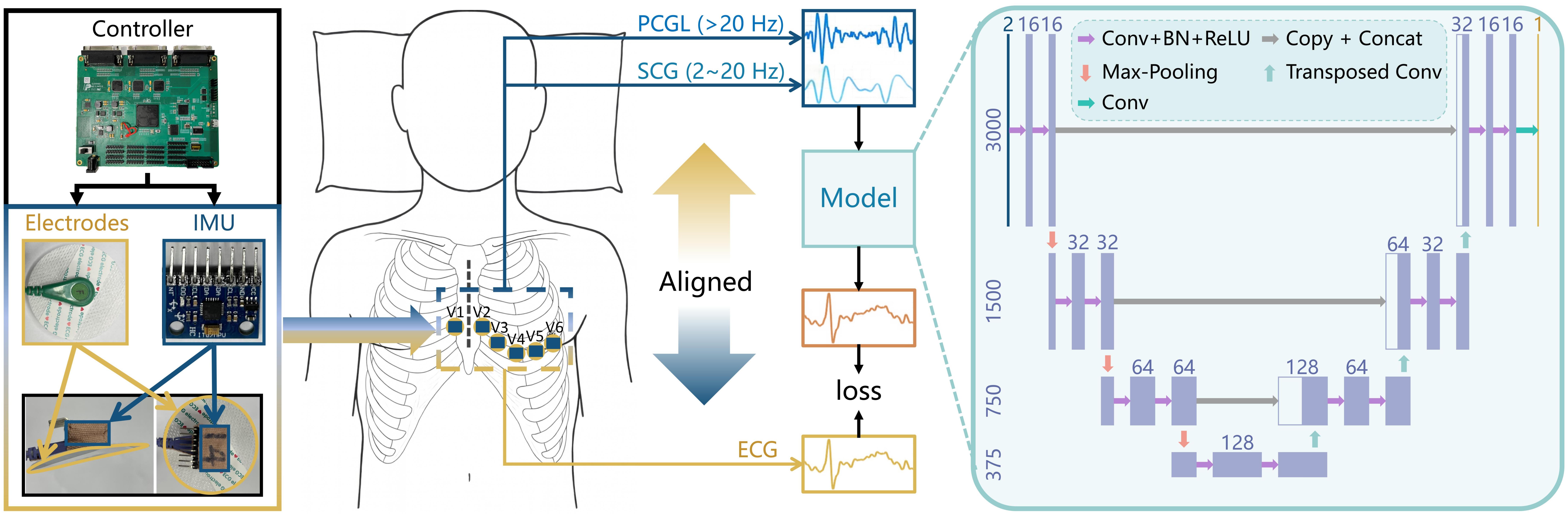}
    \caption{The overview of the model and data collection approach. 
    Left: In the hardware setup, an IMU is attached to an electrode buckle to form an electromechanical sensing pair that records colocated ECG and vibrational signals at six positions (V1–V6). All pairs are controlled by a single FPGA with a shared global clock, enabling timestamp-based interpolation for cross-modal alignment. 
    Middle: The acquired vibration signals are decomposed into low-frequency signals (SCG, 2–20 Hz) and high-frequency phonocardiogram-like components (PCGL, $>$20 Hz), which are normalized separately and then together used as model inputs.
    Right: The baseline U-Net model takes SCG and PCGL as inputs to reconstruct the corresponding ECG signals, supervised by the aligned ground-truth ECG through a reconstruction loss. }
    \label{fig:overview}
    \vspace{-3mm}
\end{figure*}
%-----------------------------------------------------------------
\section{Related Work}

To leverage both the theoretical foundation of ECG and the low cost of SCG while fully exploiting SCG information, it is feasible to collect SCG signals and convert them into ECG, due to the tight electromechanical coupling of cardiac signals~\cite{pfeiffer2014biomechanics}. \cite{park2020heartquake} mounted a geophone on a bed to collect cardiac vibration signals and used a Bi-directional Long Short-Term Memory (Bi-LSTM) to convert them into ECG signals. 
Physicians were invited to identify the signals, and except for one participant, the rest could not distinguish between real and generated signals. \cite{satija2022unified} used the CEBS open-source dataset and an attentive cycle-generative adversarial network (CycleGAN) for conversion, achieving a root mean square error (RMSE) of 0.418. \cite{cao2022guard} collected pulse signals using a wrist-worn accelerometer and converted them into ECG signals with a network combining GAN and encoder-decoder models, achieving a waveform amplitude estimation error of 5.989\%. \cite{chen2022contactless} acquired chest vibration signals non-contactly using millimeter-wave radar and performed domain transformation with a transformer-based architecture, obtaining an RMSE of 0.081 mV, a median temporal error of 14\,ms, and a median Pearson correlation of 90\% with the ground truth waveform morphology.

Despite these advances, these studies have limitations. First, based on the manuals of their devices or dataset descriptions\footnote{\cite{park2020heartquake} used the BioHarness 3.0 device to collect ECG data, which, according to its \href{https://www.zephyranywhere.com/media/download/bioharness3-user-manual.pdf}{manual}
, records only one ECG channel. \cite{satija2022unified} used the CEBS dataset containing only lead I and lead II, both limb leads. \cite{cao2022guard} used the AD8232 ECG front end \href{https://www.analog.com/en/products/ad8232.html}{data sheet}
, which supports a single lead. \cite{chen2022contactless} used the TI ADS1292 evaluation board \href{https://www.ti.com/tool/ADS1292ECG-FE\#tech-docs}{manual}
, which has two channels. Measuring one chest lead requires at least four channels.}, 
these studies are restricted to reconstructing ECG signals on limb leads from SCG collected at a fixed, single location. 
However, chest leads are essential for many cardiovascular diseases, such as myocardial infarction~\cite{namdar2018st}, ventricular fibrillation~\cite{nakano2010spontaneous}, congenital heart disease~\cite{white2025diagnostic}, ventricular tachycardia~\cite{reithmann2019electrocardiographic}, and premature ventricular contraction~\cite{yamada2019twelve}. 
Second, these studies do not provide mechanical recordings co-located with each of the six standard precordial electrodes (V1–V6). 
Finally, none of these studies examined the temporal generalization of their models. Previous research~\cite{cano2023improved} has indicated that it is unrealistic to expect a model trained on physiological data from a single day to perform well over an extended period, highlighting the need to investigate model performance over time, which requires multi-day data collection from the same subjects, a type of dataset that is currently lacking in this field.

Besides SCG, there are also studies using photoplethysmography (PPG) to reconstruct ECG~\cite{tang2023ppg2ecgps}. However, the cost of PPG is not substantially lower than that of wearable ECG devices, limiting their advantage of cost-effectiveness for long-term ECG monitoring.

In summary, Vib2ECG provides synchronized twelve-lead ECG and vibrational recordings with an IMU co-located with each of the six standard precordial electrodes (V1–V6), enabling investigation of location-dependent SCG–ECG relationships.

%-----------------------------------------------------------------
\section{Methods}
%-----------------------------------------------------------------

An overview of the model and data collection approach is illustrated in Fig.~\ref{fig:overview}. This section first describes the hardware setup for data acquisiton and the development of the dataset, followed by an introduction to data preprocessing, with a focus on the frequency-based decomposition method and its motivation from physiological and physical perspectives. Finally, the architecture of the benchmark model is presented.

\begin{figure}[tbp]
    \centering
    \includegraphics[width=0.45\textwidth]{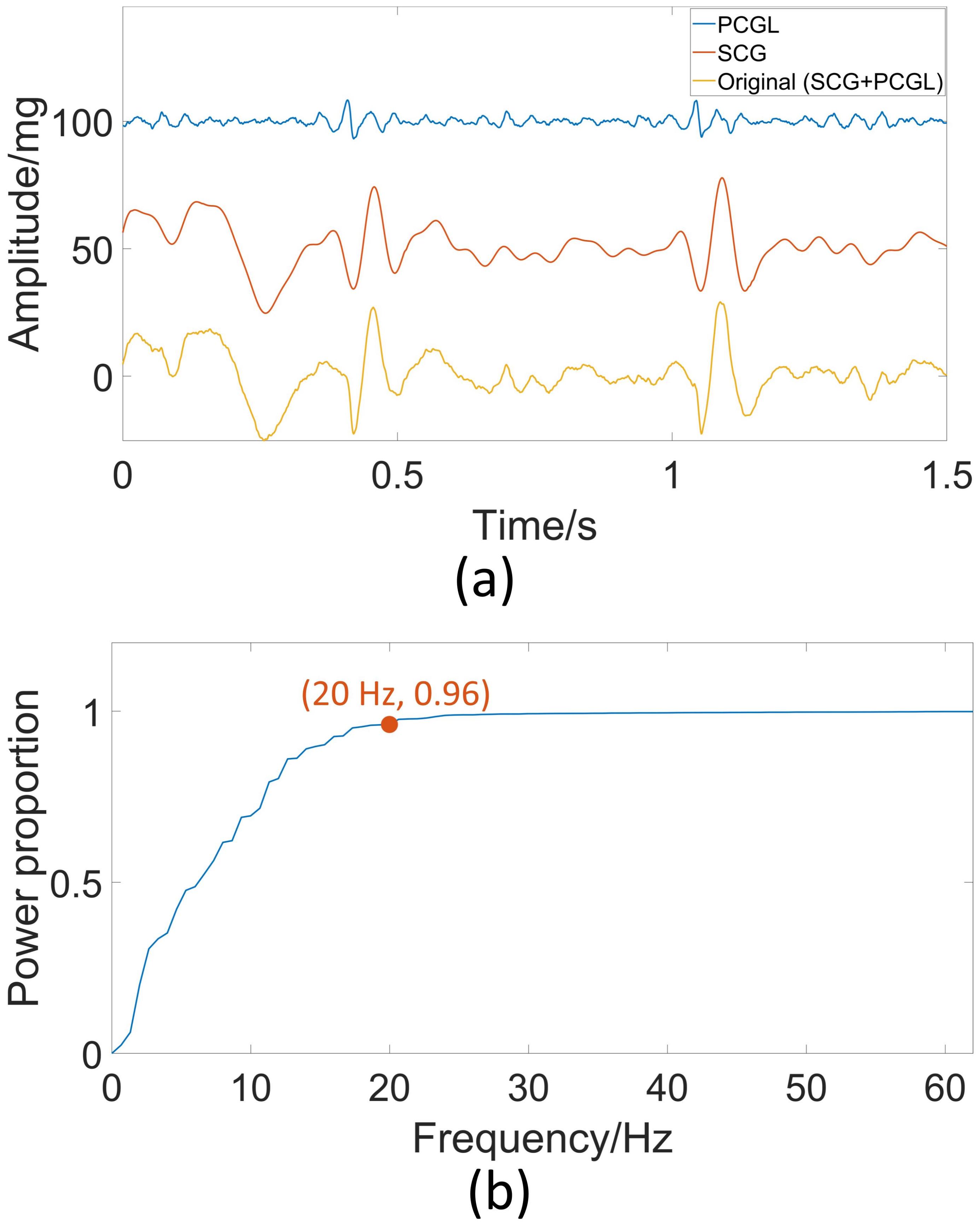}
    \caption{SCG (0-20 Hz) contains over 96\% of the total energy of the original signal. If the PCG-like (PCGL) component (above 20 Hz) is not separately extracted and normalized, it will be invisible within the original waveform, and the information carried in the PCGL band will be ignored.}
    \label{fig:SCG_PCGL_comparison}
    \vspace{-3mm}
\end{figure}

\subsection{Dataset and Hardware Setup for Data Acquisition}

As a preliminary feasibility study, this work focuses primarily on healthy participants. Data were collected from 23 healthy male subjects aged between 21 and 70 years.
However, data from 6 participants were excluded due to severe electromagnetic interference near their acquisition sites, which caused excessive noise in their chest-lead ECG signals. 
Although the corresponding limb-lead ECG signals remained relatively clean, these data were not suitable for the objectives of this work and were therefore excluded from subsequent experiments. As a result, data from the remaining 17 participants with higher-quality precordial ECG and SCG signals were used for analysis.
%%so their data were retained for potential future studies on the relationships between SCG and limb lead ECG signals or ECG denoising. 

Details of the signal acquisition system can be found in~\cite{lu2024dual}. 
The system is capable of simultaneously recording SCG and ECG signals from 16 locations in the chest. 
In this study, 6 positions were used to cover the chest leads of the standard twelve-lead ECG system. At each position, an electrode was used to collect ECG signals, while an IMU for SCG acquisition was mounted directly above the electrode via the lead wire, providing spatial co-location of each IMU with the corresponding precordial electrode. 
Both IMUs and analog-to-digital converters (ADCs)  used for ECG acquisition were controlled by a field-programmable gate array (FPGA), with all device drivers sharing a global 50\,MHz clock. 
This shared clock enabled precise cross-modal temporal alignment of ECG and SCG signals through timestamp-based interpolation, with an error below 20\,ns.

During data collection, each participant was asked to lie supine on a bed. Each experimental session lasted 1.5 hours, with 1 hour dedicated to data recording. 
Participants were instructed to remain still and avoid speaking to minimize motion-induced interference in the SCG signals. 
Due to the duration of the experiment, most participants fell asleep during recording, and occasional snoring introduced additional noise into the SCG signals. In a few cases, brief speech occurred for unavoidable reasons, which may also have introduced transient motion artifacts.

\begin{figure*}[t]
    \centering
    \includegraphics[width=\textwidth]{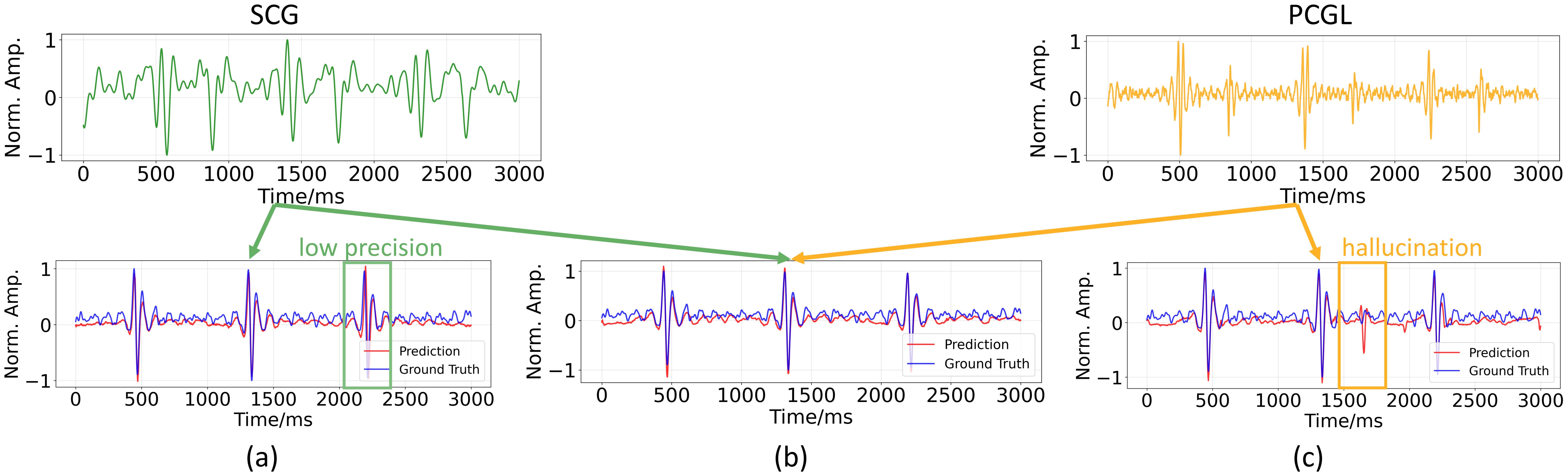}
    \caption{A typical comparative example with different inputs. (a) Reconstruction with only SCG as input. The outputs are less precise in amplitude and timing but exhibit fewer hallucinations, i.e., generated ECG waveforms that do not actually exist. (b) Reconstruction with SCG and PCGL as inputs. The outputs are precise and exhibit fewer hallucinations. (c) Reconstruction with only PCGL as input. The outputs are more precise but exhibit more hallucinations. }
    \label{fig:comparison_with_different_inputs}
    \vspace{-3mm}
\end{figure*}

\subsection{Data Preprocessing}

In the data preprocessing pipeline, ECG signals were processed using a basic high-pass filter to remove baseline drift and a low-pass filter to eliminate power-line interference and its harmonics. Vibrational signals were similarly high-pass filtered to reduce motion artifacts.

Beyond noise removal, the vibrational signals were further decomposed into low-frequency SCG ($< 20$ Hz) and high-frequency Phonocardiography-like (PCGL, $\geq 20$ Hz) components. The term “PCGL” is used because the waveform of the high-frequency component resembles that of phonocardiography (PCG)~\cite{khosrow2014automatic}. 
This separation is motivated by the complementary physiological information carried by the two components: 
SCG primarily reflects myocardial motion~\cite{rai2021comprehensive}, while PCG is more sensitive to valve motion~\cite{bhoi2015multidimensional}. The cutoff frequency of 20 Hz was chosen because PCG is essentially an acoustic signal, and 20 Hz corresponds to the lower limit of human hearing, therefore, vibrational components above this frequency are treated as sound-related signals.

The necessity of separating SCG and PCGL is illustrated in Fig.~\ref{fig:SCG_PCGL_comparison}. 
As shown in Fig.~\ref{fig:SCG_PCGL_comparison}(a), the amplitude of PCGL is smaller than that of SCG, and the original signal, which contains both SCG and PCGL, closely resembles the SCG waveform. This indicates that if PCGL is not separately extracted and normalized, feeding the original signal directly into a model is nearly equivalent to using SCG alone, and the information carried by PCGL will be ignored. Fig.~\ref{fig:SCG_PCGL_comparison}(b) further quantifies this observation: after removing the direct current (DC) component, the energy of the vibrational signal above 20 Hz accounts for less than 4\% of the total energy, further highlighting the necessity of extracting and normalizing the high-frequency component.

After the noise removal and vibrational signal decomposition, we align the vibrational signals with the ECG signals through interpolation. 
Considering hardware constraints, communication rate, and signal-to-noise ratio, the ECG ADC sampling rate is set to a higher value of 1000 Hz, while the IMU sampling rate is set to 500 Hz, resulting in misaligned sampling instances between the two modalities. However, as described in the previous subsection, the ADC and IMU drivers share the same clock on the FPGA, and both modalities transmit their data together with the corresponding timestamps to the upper computer. This allows interpolation-based alignment of the two modalities using the timestamps. In this study, taking the ECG timestamps as the reference, the SCG and PCGL signals are interpolated to 1000 Hz for alignment.

\subsection{Model}

% \begin{figure}[tbp]
%     \centering
%     \includegraphics[width=0.2\textwidth]{fig/hallucination_trend.png}
%     \caption{Variation of hallucination count with time interval between training and testing data.}
%     \label{fig:hallucination_trend}
%     \vspace{-3mm}
% \end{figure}

Our baseline model is constructed based on Wave U-Net~\cite{stoller2018wave}, which was originally developed for audio source separation. The architecture and its variants have subsequently been shown to be effective for physiological signal reconstruction~\cite{tang2023ppg2ecgps, xiang2023ecg, cheng2021prediction, pinto2024inferring}, making it suitable as a baseline model for validating the feasibility of the ECG reconstruction task in this study. 
In addition, compared with other architectures directly used for SCG-to-ECG conversion, such as GANs and Transformers~\cite{cao2022guard, satija2022unified, chen2022contactless}, Wave U-Net is more lightweight. Compared with LSTM-based models~\cite{park2020heartquake}, Wave U-Net, as a fully convolutional network, enables efficient parallel computation and results in lower computational latency~\cite{bai2018empirical}. These characteristics make Wave U-Net more suitable for the target application scenario of this study, namely, ECG reconstruction on low-cost mobile devices.

The model architecture is shown in Fig.~\ref{fig:overview}. The network adopts an encoder–decoder structure with symmetric skip connections. The input consists of two channels, corresponding to the low-frequency SCG (2–20 Hz) and high-frequency PCGL ($>$20 Hz) signals, each with a length of 3000 samples (3 s).

In the encoder, each layer contains two convolutional blocks, each composed
of a 1D convolution with a kernel size of 7, batch normalization, and ReLU activation, followed by down-sampling using max-pooling with a factor of 2. The number of feature channels progressively increases as the network goes deeper, eventually reaching 128.

In the decoder, the temporal resolution is restored using transposed convolutions, and the corresponding encoder features are concatenated via skip connections (copy-and-concatenate) to preserve fine-grained temporal information. Each layer of the decoder has the same convolutional blocks as the encoder. In the last layer, a convolution with a kernel size of 1 is used for feature fusion to generate the final prediction of ECG signals.

%The input consists of SCG and PCGL signals with a length of 3000 samples (i.e., 3 seconds), resulting in two input channels. 
%After entering the network, each layer passes through two blocks, each composed of convolution, batch normalization, and an activation function. A dropout layer with a probability of 0.1 is inserted between the two blocks in each layer. In this study, the kernel sizes for convolution, max-pooling, and transposed convolution are 7, 2, and 2, respectively. \textcolor{red}{ this paragragh is not very clear to me. the description of network needs to be reconstructed, for example, in the figure, mp, transposed conv, copy and concat, etc are mentioned but cannot find in text.}

\section{Experimental Evaluation and Results}

In this section, we first describe the experimental setup and evaluation metrics, then present an ablation study on input decomposition and experiment results on temporal generalization, and finally analyze the causes of hallucination based on these results.

\begin{figure*}[htp]
    \centering
    \includegraphics[width=\textwidth]{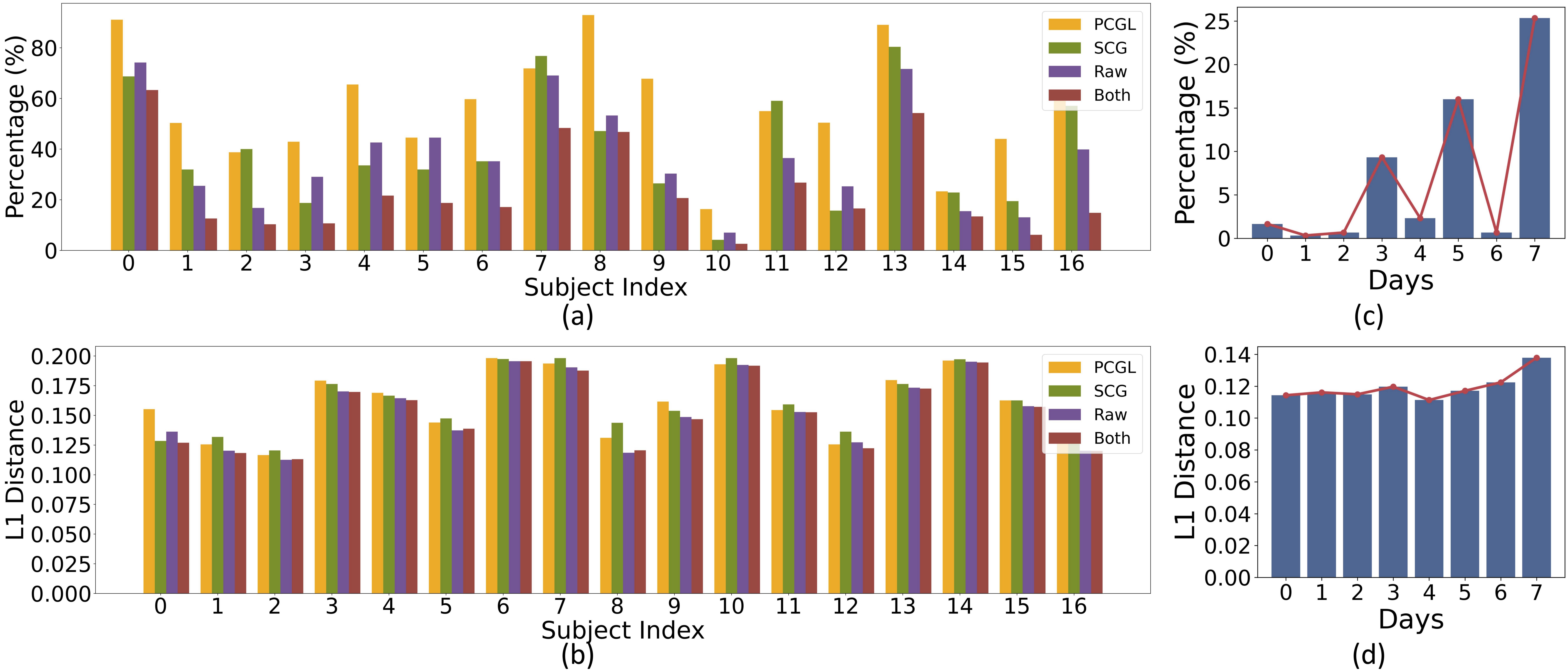}
    \caption{(a) Hallucination percentage with different inputs. “PCGL” and “SCG” denote models using a single normalized PCGL or SCG channel as input, respectively. “Raw” indicates the use of a single channel of the original acceleration signal without decomposition. “Both” represents a two-channel input consisting of normalized PCGL and SCG signals. (b) L1 distance with different inputs. (c) Hallucination percentage with different time interval between training and testing data. (d) L1 distance with different time interval between training and testing data.}
    \label{fig:comparison_and_temporal_trend}
    \vspace{-3mm}
\end{figure*}

\subsection{Experimental Setup for Model Training and Testing}\label{sec:experimental_setup}

In this study, training and testing were conducted in a subject-specific approach. For each subject's (participant's) data, the dataset was divided based on time, with the earliest 70\% of the data used for training, the middle 10\% for validation, and the last 20\% for testing. The model that achieved the best performance on the validation set was selected for testing to prevent data leakage. 

The utilized model accuracy metric is the L1 distance, because compared with other metrics such as correlation-based loss or dynamic time warping, it imposes stricter requirements on both amplitude and temporal precision, while also being more intuitive and easier to interpret. In addition, to quantify the severity of model hallucination, we randomly selected 310 reconstructed ECG signals from the test results and manually counted the occurrences of hallucinations. The total number of reconstructed ECG signals in each test result ranged from 1200 to 1450.

\subsection{Ablation Study on Input Decomposition }%\textcolor{red}{too general, option: "Ablation Study on Frequency-Based Input Decomposition", etc.}}

% \textcolor{red}{Non-decomposed input (original IMU input) as a single input channel is not evaluated quantitatively, different from figure 2, which is more intuitively explained. }

In this section, we present the model accuracy with different inputs. As shown in Fig.~\ref{fig:comparison_with_different_inputs}, comparing (a), (b), and (c) reveals that when the input contains only an SCG signal, the output exhibits lower precision in amplitude and timing but shows a lower likelihood of hallucination, i.e., generate a fake ECG waveform that does not exist. When the input contains only a PCGL signal, the output becomes more precise but hallucination occurs more frequently. When the input contains both SCG and PCGL signals, the accuracy is improved and the hallucination is diminished.

%The quantitative results are shown in Fig.~\ref{fig:comparison_and_temporal_trend}(a) and (b). In Fig.~\ref{fig:comparison_and_temporal_trend}(a), for most subjects, the hallucination frequency is highest when only PCGL is used as input, followed by the case where only SCG is used. In Fig.~\ref{fig:comparison_and_temporal_trend}(b), for most subjects, the L1 distance is largest when only SCG is used as input, indicating lower precision. Both metrics are minimized when PCGL and SCG are used jointly as input, indicating that both high-frequency and low-frequency components in the vibration signals are informative.

The quantitative results are shown in Fig.~\ref{fig:comparison_and_temporal_trend}(a) and (b). In Fig.~\ref{fig:comparison_and_temporal_trend}(a), for most subjects, the hallucination frequency is the highest when only PCGL is used as input, followed by the case where only SCG is used. In Fig.~\ref{fig:comparison_and_temporal_trend}(b), for most subjects, the average L1 distance for a subject is the largest when only SCG is used as input, indicating lower precision. In contrast, when normalized PCGL and SCG are jointly used as two input channels (denoted as “Both”), both the hallucination frequency and the prediction error are minimized, indicating that both high-frequency and low-frequency components are informative to this task.

In Fig.~\ref{fig:comparison_and_temporal_trend}(a) and (b), we also include a comparison using the raw signal as input (denoted as “Raw”), i.e., the original acceleration signal without decomposition. For most subjects, in terms of hallucination frequency, the performance of the raw signal is close to that of SCG and substantially worse than that of “Both”. In terms of precision, the raw signal performs slightly worse than “Both”. This further demonstrates that frequency-based decomposition and normalization of the input signals improve the model performance.

It is worth mentioning that in this experiment, switching the input from a single channel to two channels only changes the number of parameters in the first convolutional layer. Specifically, the model with two input channels adds only $\left(2-1\right) \times 16 \times 7 = 112$ parameters compared to the single-channel model. This increase is negligible relative to the total model size of 364\,K parameters. Therefore, the performance improvement can be attributed entirely to the change in input.

\subsection{Temporal Generalization Experiment}
\begin{figure*}[t]
    \centering
    \includegraphics[width=\textwidth]{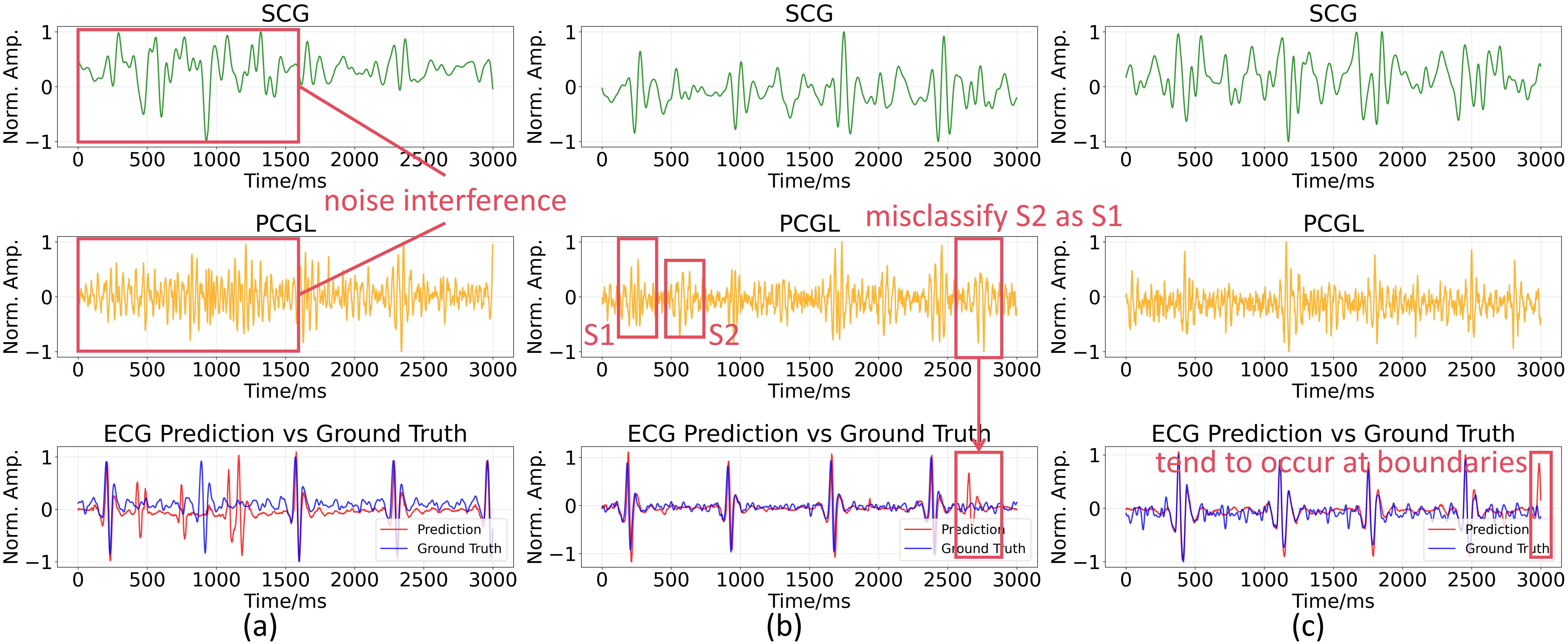}
    \caption{Factors contributing to the occurrence of hallucinations: (a) noise interference, (b) misclassification of the second heart sound (S2) as the first heart sound (S1), and (c) semantic loss at segment boundaries.}
    \label{fig:hallucination_reasons}
    \vspace{-3mm}
\end{figure*}

Intuitively, it is impractical to expect a model trained on data collected on a single day to remain effective over a long period. Therefore, unlike previous studies, in this work, the temporal generalization of the trained model is investigated. A subject was selected and data were collected in daily recording sessions over eight consecutive days. Each recording session was conducted at approximately the same time of day, resulting in an interval of about 24 hours between data from consecutive days. In this experiment, the data from the first day were used for training and testing following the procedure described in Section~\ref{sec:experimental_setup}, while the data from each subsequent day were used only for testing.

The experimental results are shown in Fig.~\ref{fig:comparison_and_temporal_trend}(c) and (d). It can be observed that both the hallucination frequency and the L1 distance exhibit oscillatory trends over time rather than monotonic increases or decreases. This suggests that, over a short-term period (one week), the occurrence of hallucinations is not dominated by the temporal gap between the training and testing data. In particular, the test results on the third, fifth, and seventh days show a noticeably higher number of hallucinations. Based on an inspection of the data from these days, we attribute the increased hallucination frequency mainly to noise, such as involuntary body movements, coughing or vocalization by the subject.

\subsection{Causes of Hallucination}\label{sec:cause_of_hallucination}

By analyzing the results of the experiments in the previous two sections, three main categories of causes can be identified: noise interference, misclassification of the second heart sound (S2) as the first heart sound (S1), and semantic loss at segment boundaries. Typical examples are illustrated in Fig.~\ref{fig:hallucination_reasons}.

Noise interference is the primary cause of hallucinations. Noise obscures the waveforms corresponding to cardiac systole and diastole in SCG and PCGL signals, making it difficult to determine when, or even whether, a heartbeat has occurred, thereby inducing hallucinations. 
In addition to the factors mentioned in the previous subsection, such as body movement, coughing, and vocalization, the primary source of noise is snoring. For subjects with relatively severe snoring, namely Subjects 7, 8, and 13, Fig.~\ref{fig:comparison_and_temporal_trend}(a) shows that their hallucination frequencies are at least 20\% higher than those of other subjects. It is worth noting Subject 0, whose data are not affected by snoring but still exhibit a higher hallucination frequency. A possible explanation is that this subject has relatively weak cardiac contractions combined with thicker subcutaneous fat, resulting in vibration signals with smaller amplitudes that are difficult for the model to capture, thereby leading to a high hallucination frequency.

Misclassification of the second heart (S2) sound as the first heart (S1) sound can explain why using only PCGL as input produces more hallucinations than using only SCG. S1 mainly originates from vibrations caused by the closure of the mitral and tricuspid valves and marks the onset of ventricular systole, whereas S2 mainly originates from vibrations caused by the closure of the aortic and pulmonary valves and marks the onset of ventricular diastole. In PCGL signals, the waveforms of S1 and S2 are similar in shape, with the primary difference being that the amplitude of S1 is generally larger than that of S2. However, this distinction is not reliable. For nearly every subject, there exist cardiac cycles in which the amplitude of S2 approaches that of S1, or cases where, although the S2 amplitude is smaller than S1 within the same cycle, it is comparable to or even exceeds the S1 amplitude in other cardiac cycles. In contrast, SCG waveforms during systole and diastole differ in both shape and amplitude and are less prone to confusion. Therefore, in the absence of SCG as a reference input, the similarity between S1 and S2 in PCGL can confuse the model, causing it to generate an ECG waveform during diastole that should have appeared during systole. Moreover, hallucinations induced by this mechanism tend to have more complete waveforms and larger amplitudes compared with those caused by noise.

In addition to the two causes discussed above, hallucinations also tend to occur at segment boundaries. We assume that this is because the model cannot observe complete waveforms and sufficient contextual information at the boundaries, leading to errors in determining whether a heartbeat is present.

\section{Discussion}

This section first discusses possible directions for mitigating model hallucination, then presents insights gained from this study, including the significance of high-frequency components in vibration signals and the potential for deployment on mobile devices, and finally explores other possible applications of the dataset.

\subsection{Potential Directions for Mitigating Model Hallucination}\label{sec:solution_of_hallucination}

\subsubsection{Improving Input Signal Quality}
\begin{figure}[tbp]
    \centering
    \includegraphics[width=0.45\textwidth]{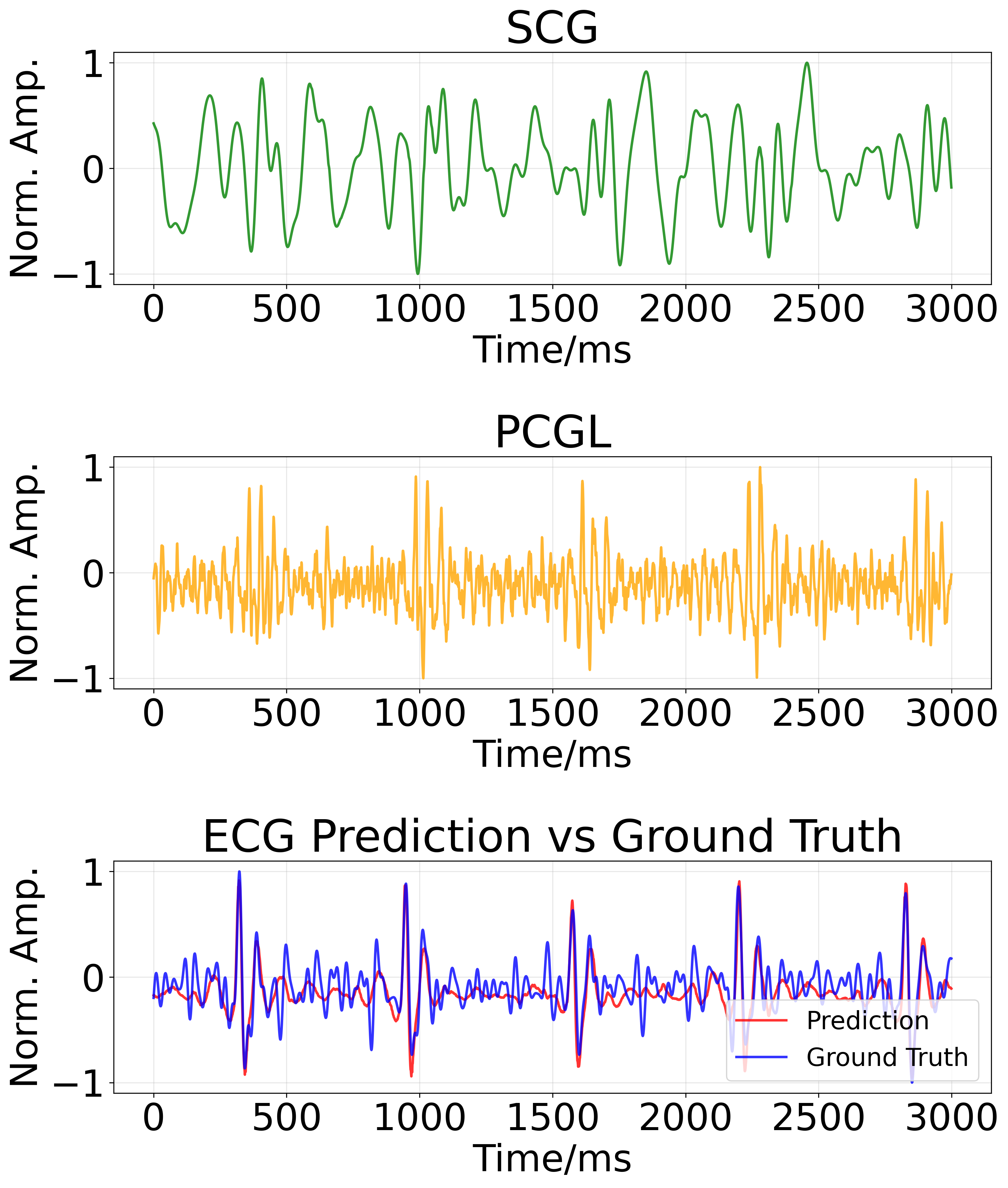}
    \caption{Results from Subject 10 demonstrate that as long as the input signal quality is sufficient for the model to obtain enough information to determine whether and when a heartbeat occurs, no hallucinations are produced in the output, even when the ECG signal used as ground truth contains substantial noise.}
    \label{fig:hallucination_depend_on_inputs}
    \vspace{-3mm}
\end{figure}

Improving the input signal quality is the most direct way to mitigate hallucinations. As explained in Section~\ref{sec:cause_of_hallucination}, when the input signal quality is sufficient to provide reliable information indicating cardiac systole and diastole, the hallucination frequency decreases significantly. In addition, our observations indicate that the occurrence of hallucinations is not strongly correlated with the training data label quality, i.e., the noise level of the ECG reference signal. Fig.~\ref{fig:hallucination_depend_on_inputs} shows a representative data segment from Subject 10. In this subject’s data, the ECG signal (the blue curve in the figure) contains relatively substantial noise, which results in a relatively large L1 distance compared with most other subjects, as shown in Fig.~\ref{fig:comparison_and_temporal_trend}(b). However, the input SCG and PCGL signal quality for this subject is relatively high—specifically, at least one input channel exhibits clearly discernible periodic waveforms—leading to the lowest hallucination frequency among all subjects, as shown in Fig.~\ref{fig:comparison_and_temporal_trend}(a). The testing results from Subject 10 further emphasize that improving the input data quality can effectively reduce the hallucination frequency, even when the label quality is relatively poor.

\subsubsection{Correlate R-peak and Heart Sounds}
A second major contributor to hallucinations is confusion between S2 and S1, which is particularly pronounced when using only PCGL as input.
Compared with noise-induced hallucinations, these cases are often easier to identify because they tend to occur temporally close to S2 and far from the expected timing of S1, and they frequently resemble true QRS complexes with larger and more complete waveforms.
%Hallucinations caused by this factor are relatively easy to identify, as they tend to occur close to S2 and far from S1. Compared with noise-induced hallucinations, their waveforms are usually larger and more complete, and more closely resemble true QRS complexes. 
A practical mitigation strategy is to enforce a physiological consistency check between heart sounds and predicted R-peaks.
Specifically, S2 can be detected from the input PCGL signal, while QRS complex or R-peak detection is applied to the reconstructed ECG.
Hallucinations can then be flagged when an R-peak occurs in an implausible temporal relationship with detected heart sounds (e.g., abnormally close to S2 or outside an expected timing window).
%In principle, a workflow can be constructed in which second heart sound detection is performed on the input PCGL signal and R-peak detection is performed on the output signal, and the distance between the two is compared to determine whether a hallucination has occurred based on a predefined threshold. 
The key components of this strategy, namely QRS detection and heart sound detection, have been extensively studied and are supported by mature algorithms, such as those in~\cite{pan1985real} and~\cite{springer2015logistic}. Therefore, this issue is theoretically solvable.

\subsection{Significance of High-Frequency Components}

In signal processing, smoother signals are often intuitively preferred, and waveform “spikes” are typically regarded as undesirable, which often leads to the removal of high-frequency components. SCG signals are no exception, and previous studies, such as~\cite{park2020heartquake, cao2022guard}, explicitly retained only the frequency range between 5 and 30~Hz. Only a limited number of works have explicitly noted that high-frequency components may serve as a useful complement to SCG signals~\cite{khosrow2014automatic}. In our study, the ablation analysis provides further evidence of the potential value of high-frequency components, namely PCGL, highlighting that components often regarded as noise may provide informative clues for SCG-related tasks.

\subsection{Potential of Applications on Mobile Devices}

Our results suggest that simultaneously and synchronously acquired ECG and vibrational signals (including SCG and PCGL) at the same chest location provide strong temporal and physiological alignment, which helps the model learn more cross-modal correlated features, without relying on structurally complex or computationally intensive model architectures. Compared with previous SCG-to-ECG studies using models such as GAN, Transformer, or Bi-LSTM~\cite{satija2022unified, cao2022guard, chen2022contactless, park2020heartquake}, this work employs a lightweight Wave U-Net with only 364\,K parameters, making it more suitable for resource-constrained and mobile-device scenarios. 
This design also aligns with the intended use case: placing a mobile device with an IMU (e.g. a smartphone) on the chest to directly reconstruct the corresponding chest-lead ECG signal. Compared with configurations limited to one or two fixed limb leads in prior studies, the proposed lead configuration has the potential to offer improved clinical relevance, enhanced flexibility, and value, while enabling low-cost, convenient, on-device ECG monitoring in practical settings. 

%At the same time, this approach of converting signals collected synchronously at the same location aligns with our practical application goal, namely placing a mobile device with an IMU, such as a smartphone, on the chest to directly output the corresponding chest-lead ECG signal, which has higher clinical value in medical practice.

\subsection{Other Potential Uses of the Dataset}

In addition to build a model for reconstruction of ECG from SCG signals, our dataset can also be used for other purposes. Potential applications are described as follows:

\subsubsection{Electromechanical Coupling Analysis}
As ECG and SCG signals in our dataset are collected synchronously at the chest-lead positions of the standard twelve-lead ECG system, the dataset provides an opportunity to investigate the electromechanical coupling relation between SCG and ECG and the corresponding physiological activities from a medical perspective, thus provides support for SCG interpretation.

\subsubsection{Optimal Position for Single-Channel Wearable SCG}
For wearable SCG applications that rely on a single sensor, choosing an optimal placement location is critical, as signal quality and waveform morphology can vary substantially across chest positions.
Previous studies typically repeated experiments by placing a single sensor at different positions sequentially, which may introduce confounding effects due to subject fatigue, posture changes, or adaptation across repeated trials and thus affect experimental results. 
In contrast, our dataset provides SCG signals recorded synchronously at multiple chest locations, enabling fair comparisons of candidate placements under matched physiological and behavioral conditions.
%%In contrast, our dataset contains SCG signals synchronously collected at multiple positions, enabling the study of optimal SCG acquisition locations while eliminating this confounding factor, and thus has the potential to provide useful insights for related research.

\subsubsection{Denoising}
Long-term cardiac monitoring studies, particularly during sleep, inevitably involve sleep-related artifacts such as snoring and involuntary body movements. 
During the data collection process of this dataset, most subjects fell asleep, resulting in recordings that naturally contain sleep-related noises. Therefore, this dataset can support research on SCG signal denoising, thereby facilitating long-term cardiac monitoring, particularly for sleep-related cardiac monitoring applications scenarios.

\section{Limitations and Future Work}

\subsection{Limitations}

As an exploratory study, this work has several limitations. First, being constrained by the equipment and experimental environment, the overall quality of the ECG signals used as ground truth is not ideal.
This primarily affects the precision of the quantitative performance evaluation, but does not undermine the feasibility of the proposed cross-modal reconstruction framework.
Second, commonly used metrics in prior studies, such as correlation-based loss or dynamic time warping loss, are not sensitive to the hallucination phenomenon identified in this work. 
As a result, manual inspection and counting were adopted to estimate the hallucination frequency. Although effective, this approach is time-consuming and may introduce subjective bias.
%As a result, manual counting was adopted in this study to quantify the severity of hallucinations. However, this approach is inefficient and can introduce subjective bias, affecting the estimation of hallucination frequency. 
Third, the temporal generalization experiment was conducted on data from a single subject due to the high cost and complexity of long-term data collection. 
While the results provide preliminary insights, the limited sample size prevents more general conclusions regarding long-term robustness.

\subsection{Future Work}

This study demonstrates the feasibility of converting cardiac vibrational signals into ECG signals using lightweight models. Based on this work, several directions can be further explored:

\subsubsection{Adapting Spiking Neural Networks}
As analyzed in Sections~\ref{sec:cause_of_hallucination} and~\ref{sec:solution_of_hallucination}, reconstructing ECG signals may primarily rely on information such as the timing of cardiac beats. This suggests that the informative content in SCG and PCGL signals is sparse, which is well suited for spiking neural networks (SNNs). Therefore, a potential direction is to identify critical features and encode the inputs into spikes accordingly for processing, thereby further improving computing efficiency.

\subsubsection{Vibrational Signal Denoising}
The analyses in Sections~\ref{sec:cause_of_hallucination} and~\ref{sec:solution_of_hallucination} indicate that the input signal quality is the dominant factor affecting the model performance. Accordingly, a promising direction is to develop denoising algorithms for the input signals, such as methods to suppress interference caused by snoring.

\subsubsection{Enhancing the Dataset and Evaluation Metrics}
Future work will also focus on expanding the dataset, particularly by collecting long-term recordings from multiple subjects to enable more comprehensive temporal generalization analysis.
In addition, developing objective and automatically computable metrics for hallucination evaluation is an important direction.
%In this study, the amount of data collected for the temporal generalization experiment is limited, and additional data are needed in future work. Moreover, more objective and automatically computable metrics for evaluating hallucinations should be developed to improve experimental efficiency and reduce subjective bias.

\section{Conclusion}

This work presents a dataset comprising synchronized twelve-lead ECG and vibrational recordings, with an IMU co-located with each of the six standard precordial electrodes (V1–V6). Using this dataset, we demonstrate the feasibility of training and using a lightweight neural network model to reconstruct corresponding ECG signals from cardiac vibrational signals at variable locations through experiments. We further analyze the hallucination phenomenon of the model, identify its primary causes, and provide potential mitigation strategies.

The low acquisition cost of acceleration-based signals and the widespread clinical use of the twelve-lead ECG system support the application potential of this study and the released dataset. Moreover, the lightweight U-Net model employed in this work, containing only 364\,K parameters, further suggests the feasibility of deploying such models on mobile devices without requiring additional hardware. This reveals a potential vision in which a mobile device equipped with an IMU, such as a smartphone, can be placed on the chest to automatically reconstruct the corresponding chest-lead ECG signal, significantly reducing the hardware cost while preserving the clinical relevance of the reconstructed ECG signal. This approach opens new possibilities for large-scale, long-term, and more interpretable cardiac monitoring.

\section*{References}
\bibliographystyle{IEEEtran}
\def\refname{\vadjust{\vspace*{-2.5em}}} %Please don't do this in a real paper.
\bibliography{refs.bib}

% \begin{IEEEbiography}[{\includegraphics[width=1in,height=1.25in,clip,keepaspectratio]{qinyu0.png}}]{Qinyu Chen} (Member, IEEE) received the Ph.D. degree from Nanjing University, China, in 2021. 
% She is an Assistant Professor at the Leiden Institute of Advanced Computer Science, Leiden University, the Netherlands. In 2022, she joined the Institute of Neuroinformatics, University of Zürich and ETH Zürich, Switzerland as a postdoc. 
% Her current research interests include the seamless neuromorphic AI system at the edge of healthcare, as well as AR/VR with a focus on event-based processing. In 2022, She received a Bridge Fellowship Grant from the Swiss National
% Science Foundation (SNSF). She also serves on the Neural Systems and Applications (NSA) Technical Committee in the IEEE Circuit and System Society (CASS). 
% \end{IEEEbiography}

\end{document}